# Deep Learning-Based Automated Quantification of TIMI Myocardial Perfusion Frame Count (DL-TMPFC) from Coronary Angiography: A Novel Framework for Rapid Assessment of Microvascular Dysfunction

Si Li[1], Yuanqing He[1], Chenkai Hu[2*], Xiaogang Guo[3*], Huay-Cheem Tan[4*], Chieh Yang Koo[4], Xuan Zhang[3], Lei He[2], Jingyuan Zeng[2], Shan Xiao[2]

[1]*School of Artificial Intelligence and Digital Economy Industry, Guangzhou Institute of Science and Technology, Guangzhou, China;*

[2]*Department of Cardiology, Second Affiliated Hospital, Jiangxi Medical College, Nanchang University, Nanchang, China.*

[3]*Department of Cardiology, First Affiliated Hospital of Zhejiang University School of Medicine, Hangzhou, China.*

[4]*Department of Cardiology, National University Heart Centre, Singapore.*

**Aims**

Coronary microvascular dysfunction (CMVD) affects approximately 40%-60% of patients with ischemia and non-obstructive coronary arteries, yet diagnosis remains challenging due to reliance on invasive functional testing or subjective Thrombolysis In Myocardial Infarction (TIMI) flow grade. The TIMI Myocardial Perfusion Frame Count (TMPFC) offers an objective, angiography-based quantitative measure of CMVD, but its clinical translation is hindered by cumbersome manual calculation and insufficient validation. This study aims to develop and validate a deep learning-powered TMPFC calculation (DL-TMPFC), enabling integration into clinical workflows.

**Methods and results**

DL-TMPFC framework comprised two components. A stenosis detection network first excluded obstructive coronary artery disease (CAD). A territory-aware segmentation network then identified perfusion territories and TMPFC calculation module automatically determined the first and last frames from angiographic sequences. The framework was validated in a cohort of 655 patients (445 of obstructive CAD, 100 of confirmed CMVD, 110 of control group) from three independent institutions. DL-TMPFC showed excellent agreement with expert manual measurements (bias: -0.93 frames; 95% LoA: -5.33 to +3.47; $r$ =0.98). DL-TMPFC markedly enhanced clinical feasibility by fully automating TMPFC and removing observer dependence. Clinically, DL-TMPFC accurately identified CMVD across a full spectrum of coronary pathologies and captured the continuous severity of CMVD beyond binary classification, enabling quantitative risk stratification.

**Conclusion**

DL-TMPFC enabled automatic, standardized, and accurate quantification of CMVD directly from routine angiography. By providing an automatic and objective measure, this tool provided immediate diagnostic information for timely recognition and management of CMVD in clinical practice.

## Introduction

Coronary microvascular dysfunction (CMVD) affects approximately 40%-60% of patients with ischemia and non-obstructive coronary arteries[1]. Patients with CMVD have a two to four-fold increased risk of major adverse cardiovascular events (MACE) compared with those without CMVD[1]. This excess risk is independent of epicardial coronary artery disease and contributes to recurrent hospitalizations for refractory angina and heart failure. Despite its high prevalence, CMVD is often underrecognized, as routine angiography fails to visualize the microcirculation. Thus, CMVD represents a prevalent yet frequently missed cause of myocardial ischemia.

An automatic, simple, standardized, and accurate diagnostic tool for CMVD remains an important unmet clinical need. Invasive indices such as coronary flow reserve (CFR) and index of microcirculatory resistance (IMR) are reference standards but are constrained by complexity, cost, and limited availability[2]. The TIMI Myocardial Perfusion Frame Count (TMPFC) offers an invasive, objective, angiography-based quantitative measure of microvascular function by assessing contrast transit time through the myocardium[3-4]. However, manual TMPFC calculation requires frame-by-frame identification of contrast entry and clearance. It is a labor-intensive, observer-dependent process that is impractical for real-time clinical use, preventing its widespread adoption. In routine practice, CMVD diagnosis is often inferred from qualitative TIMI flow grade assessment and clinical presentation, both subjective and imprecise.

Deep learning-based TMPFC has the potential to provide an automatic, simple, standardized, and accurate diagnostic tool for CMVD. Recent advances in deep learning have transformed cardiovascular imaging, enabling automated analysis with high precision. While prior studies have focused on stenosis detection[5] or fractional flow reserve (FFR) estimation[6], the automated quantification of TMPFC remains unexplored. By automating the entire TMPFC calculation process, deep learning eliminates observer-dependent variability and reduces analysis time from minutes to seconds, making it feasible for real-time clinical use.

This paper proposes a deep learning framework (DL-TMPFC) for fully automated TMPFC quantification. The workflow comprises stenosis detection and TMPFC calculation phase. A stenosis detection network assessed obstruction severity. Cases flagged with no significant stenosis were proceeded as potential CMVD. In the TMPFC calculation, a territory-aware segmentation network produced separate binary masks for the left anterior descending artery (LAD), left circumflex artery (LCX) and right coronary artery (RCA) perfusion territories. The initial (contrast entry) and the last frames were identified by TMPFC measurement module. The interval yielded the TMPFC, replacing error-prone manual analysis. This dual-phases approach integrates stenosis screening with perfusion quantification, optimizing efficiency and objectivity in CMVD assessment and offering a rapid preliminary assessment tool to improve CMVD diagnosis, risk stratification, and therapeutic monitoring.

# Methods

## 1. Patient enrollment and data acquisition

This retrospective analysis included 655 consecutive patients who underwent elective percutaneous coronary intervention (PCI) at the Second Affiliated Hospital of Nanchang University (Nanchang, China), the First Affiliated Hospital of Zhejiang University School of Medicine (Hangzhou, China) and the National University Heart Centre, Singapore (Singapore) between December 2023 and December 2025. Unlike the original TMPFC study[3-4] in patients with acute ST-segmentation elevation myocardial infarction (STEMI) and healthy controls, our retrospective cohort comprised symptomatic patients without acute events. This study focused on diagnostic performance rather than prognosis in original study. The patients were divided into three groups. Group A were 445 cases with obstructive CAD. Group B were 100 cases with confirmed CMVD. Group C were 110 control patients with mostly early coronary atherosclerosis. Symptom overlap among the three groups was possible, yet grouping by the primary pathology was clinically reasonable.

The study protocol received institutional ethics committee approval, and written informed consent was obtained. Exclusion criteria included clinical contraindications, acute coronary syndromes, procedural limitations (chronic total occlusions, extreme vessel tortuosity precluding intravascular imaging catheter advancement). All participants underwent conventional coronary angiography using iodixanol-320 (1-2 cc/s per territory) on a Philip Allura Xper

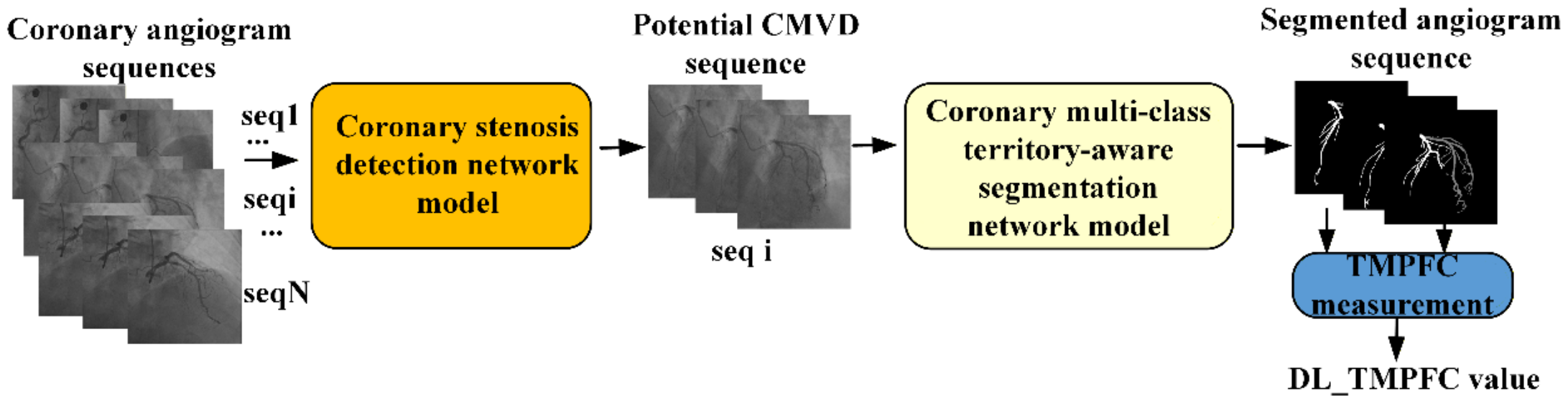

**Figure 1.** Schematic of the proposed DL-TMPFC framework.

FD20 system, with 7-10 projections per patient and more than 30 frames per sequence. To select the optimal angiogr-aphic view, it was necessary to ensure that the vessels had no severe foreshortening and that the vessel structure could be fully unfolded to facilitate observation. The LCX and LAD vessels must not overlap with each other. The recommended views were Left anterior oblique with Cranial or Right anterior oblique with Cranial for LAD, Right anterior oblique with Caudal or Left anterior oblique with Caudal for LCX, Left anterior oblique with Cranial or Right anterior oblique for RCA.

## 2. Proposed DL-TMPFC framework

The proposed DL-TMPFC framework comprised stenosis detection phase and TMPFC calculation phase. For stenosis detection network, the training set contained opacified frames with manually annotated bounding boxes. For territory-aware segmentation network, the training set contained opacified frames with manual labels. Manual annotations were performed separately to independent binary masks for LAD and LCX, and a single mask for RCA.

At inference as shown in Figure 1, raw angiograms were first processed by the stenosis detection network. Cases without more than 50% stenosis were flagged as potential CMVD and passed to the segmentation network, which generated frame-wise binary masks for the target territory. The first (contrast entry) was identified by backward search from peak opacification as the earliest frame reaching a spatially adaptive threshold of the maximum pixel count. The last frame (contrast clearance) was detected by forward search as the first frame where pixel count fell below the lowest of three thresholds. There were a fixed percentage of peak value, the mean pixel count of initial frames, and the mean of final frames. The interval between these frames gave the TMPFC, enabling second-scale automated computation. The following sections described three core components: stenosis detection network development, DL-driven TMPFC measurement, and clinical validation.

## 3. Stenosis detection network development

The stenosis detection network development comprised three steps (shown in Figure 2): manual annotation, network training, and performance validation. Cardiologists selected contrast-opacified frames from angiograms. Two experts annotated stenosis bounding boxes using LabelImg[7], focusing on critical segments, such as LAD proximal and middle segments, LCX ostial regions, and RCA middle and distal segments. Labels underwent rigorous quality control, with inter-observer variability assessed via Cohen's $\kappa$-coefficient[8] (greater than 0.81 required). This operation yielded ground-truth detection bounding boxes essential for supervised learning.

The training set consisted of 70% of patients from each of group A, B, and C, while the test set consisted of the remaining 30% from each group. We employed the YOLO12[9] network, an architecture renowned for its high precision and efficiency in real-time object detection. The network was optimized using stochastic gradient descent. Training leveraged paired inputs of the angiogram frames and corresponding expert-annotated detection bounding boxes. The optimization utilized YOLO12's composite loss, integrating Distribution Focal Loss (DFL), Complete

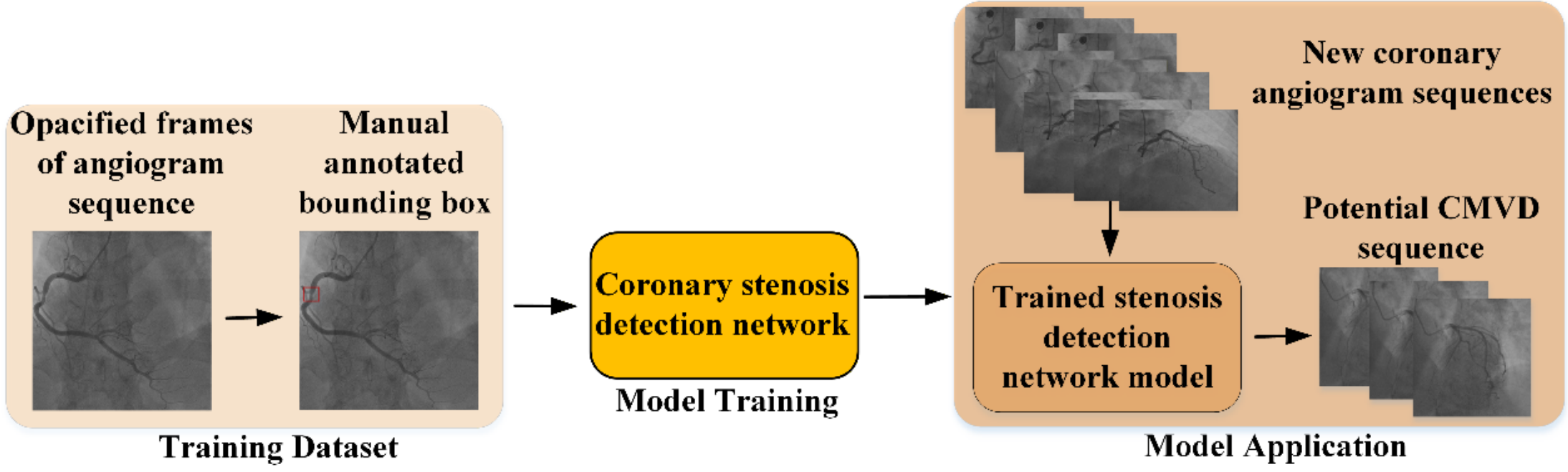

**Figure 2.** Detailed description of the development of coronary stenosis detection network model.

Intersection over Union (CIoU) Loss for regression, and a weighted Binary Cross-Entropy loss for classification tasks. Hardware acceleration (NVIDIA A10 GPUs with 24GB memory) enabled batch size of 32 across 600 epochs. The model classified stenosis as non-obstructive (stenosis severity < 50%) and obstructive (stenosis severity > 50%). Non-obstructive cases proceeded to the following DL-driven TMPFC measurement. The diagnostic accuracy on the test set was using mAP50, mAP50-95, precision and recall to evaluate the detection performance for accuracy metrics[10]. GFLOPs and params were adopted to assess model efficiency and complexity[11].

## 4. DL-driven TMPFC measurement

The DL-driven TMPFC measurement method comprised two key components: development of a coronary multi-class territory-aware segmentation network and TMPFC measurement.

### 4.1 Coronary multi-class territory-aware segmentation network

The segmentation network development comprised three steps (Figure 3(a)): manual annotation, model training, and model validation. Two cardiologists annotated coronary boundaries per vessel using the specialized software ITK-SNAP[12], creating independent masks for LAD, LCX, and RCA territories. The dataset split was identical to that used for the detection network.

We adopted the MedFormer network[13] optimized using stochastic gradient descent, using paired raw angiograms and expert masks. Data augmentation (geometric transformation and signal interference addition) enhanced robustness against clinical variability. The loss function combined Dice coefficient and cross-entropy, with early stopping to prevent overfitting[13]. Hardware acceleration (NVIDIA A10 GPUs with 24GB memory) enabled batch size of 8 across 1000 epochs.

Qualitative analysis was conducted by three independent cardiologists employing a 5-point Likert scale[14] (1 means non-diagnostic, 5 means excellent) on the test sets across three challenging contexts: low-contrast runoff territories, complex vascular geometries, indistinct boundary regions. Key evaluation criteria included vessel continuity preservation at bifurcation sites, contour accuracy within stenotic segments, and specificity for coronary anatomy.

Quantitative evaluation employed on the test dataset using metrics including Dice Similarity Coefficient (DSC), sensitivity, 95% Hausdorff distance (HD95) and Intersection over Union (IoU)[15]. The expert annotations were adopted as the benchmark.

### 4.2 TMPFC measurement

#### 4.2.1 Algorithm workflow

For each frame of the segmentation mask, preprocessing was applied (removal of border artifacts and small connected components). The number of vessel region pixels in each frame was counted as $A_t$. The peak opacificatio-n frame $F_{max}$ was identified as the frame with the maximum $A_t$. Based on the smoothed $A_t$ sequence, the initial filling frame $F_1$ and the first clearance frame $F_2$ were detected (shown in Figure 3(b)). A territory-aware strategy was adopted, allowing different $F_1$ detection schemes for different coronary branches (LCX, LAD, RCA). The TMPFC value was then calculated from $F_1$ and $F_2$.

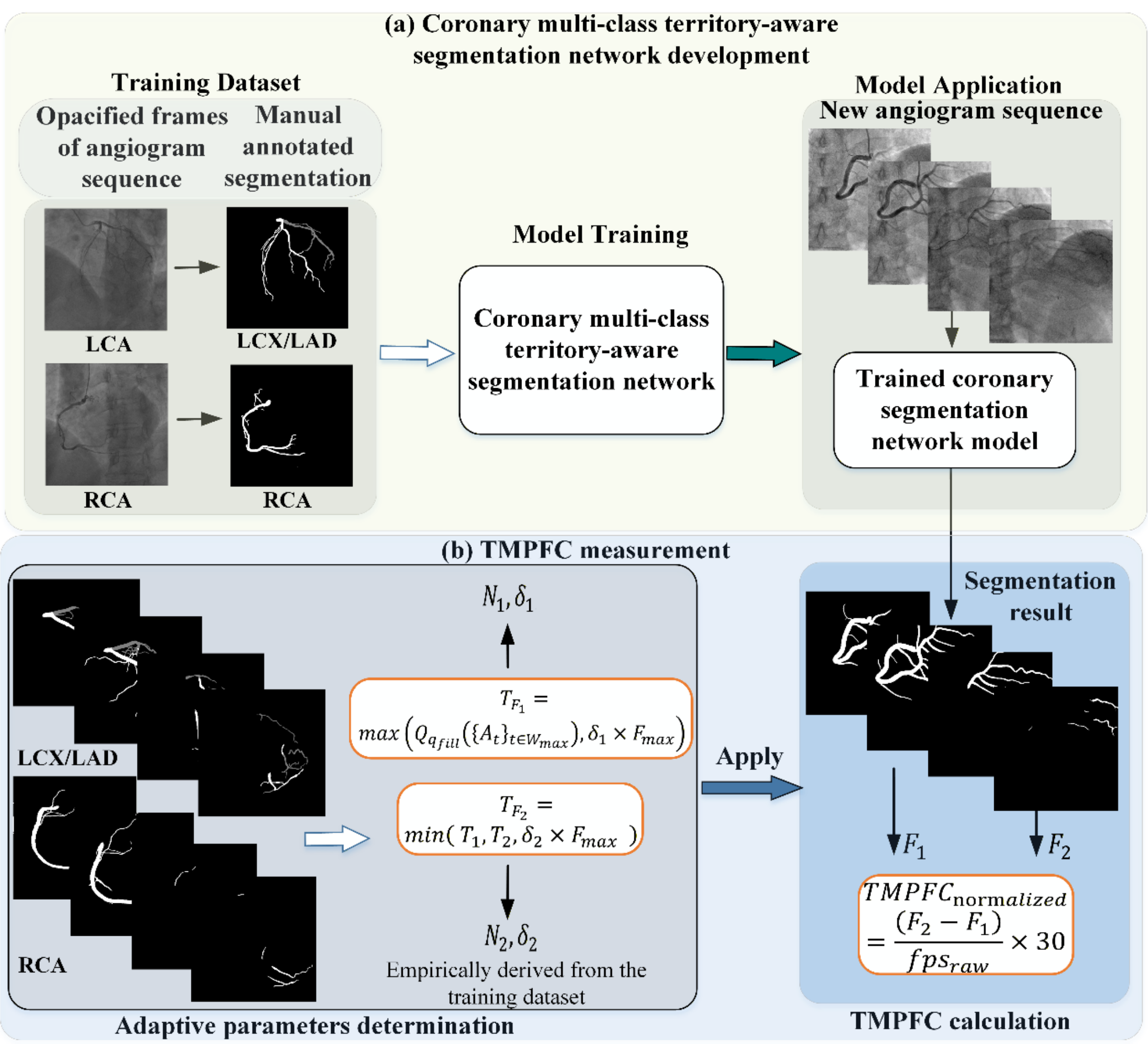

**Figure 3.** Detailed description of the development of DL-driven TMPFC measurement method.

4.2.2 $F_1$ detection (initial filling frame)

For $F_1$ detection, median filtering was first applied to the $A_t$ sequence. Then, a local near-peak window $W_{max}$ (window width $N_1$) around $F_{max}$ was extracted to compute the filling threshold $T_{F_1}$, shown in formula (1):

$$T_{F_1} = max\left(Q_{q_{fill}}(\{A_t\}_{t \in W_{max}}), \delta_1 \times F_{max}\right) \quad (1)$$

$Q_{q_{fill}}$ was the quantile of the near-peak window, $\delta_1$ was the parameter to set the lowest threshold. Starting from $F_{max}$ and scanning backward, the earliest frame within the high-filling stable segment that exceeded the threshold $T_{F_1}$ was taken as candidate $F_1$. Further confirmation was then performed to verify there was two frames sustained positive slope increase after this point, thereby preventing premature triggering.

4.2.3 $F_2$ detection (first clearance frame)

The threshold for detecting $F_2$ was defined in formula (2):

$$T_{F_2} = min(T_1, T_2, \delta_2 \times F_{max}) \quad (2)$$

$T_1, T_2$ were the quantile values of $N_2$ frames at the beginning and the end of $A_t$ sequence, respectively. $\delta_2$ was the parameter to set the lowest threshold. Staring from $F_{max}$ and scanning forward, the earliest frame within the contrast clearance stable segment that fell below the threshold $T_{F_2}$ was taken as candidate $F_2$. Further confirmation that the three subsequent frames remained continuously below the threshold.

TMPFC was computed as formula (3):

$$TMPFC_{raw} = F_2 - F_1 \quad (3)$$

The TMPFC value was normalized to the equivalent frame count at 30 frames per second (fps) based on the original rate recorded in the DICOM files to enable comparative analysis. The normalized value was computed as formula (4):

$$TMPFC_{normalized} = \frac{TMPFC_{raw}}{fps_{raw}} \times 30 \quad (4)$$

where $\mathrm{TMPFC}_{normalized}$ was the normalized value, $fps_{raw}$ meant the fps of the current acquisition rate.

Quality control excluded sequences with (i) $A_{max}$ ($A_{max} = A_{F_{max}}$) less than 800, (ii) invalid frame order ($F_1 \geq F_2$), (iii) missing $F_1$ or $F_2$ detection, (iv) TMPFC less than 10.

## 5. Comprehensive performance validation

To rigorously evaluate the clinical utility of the proposed DL-TMPFC framework, we designed a comprehensive validation study comprising three core components: 1) technical validation against expert manual measurements, 2) clinical validation against established patient phenotypes, 3) incremental value assessment.

### 5.1 Technical validation: agreement versus manual TMPFC

The sequences from 30 test cases of group B were used for this analysis. The ground truth for TMPFC was established by three independent interventional cardiologists, blinded to the DL-TMPFC results, who performed manual frame-counting according to the standard TMPFC definition.

The agreement between the proposed DL-TMPFC outputs and the manual ground truth was quantified using Bland-Altman analysis[16] to calculate the mean bias and 95% limits of agreement. Proportional bias was tested via linear regression of the differences against the means of the two values. Additionally, Pearson's correlation coefficient[17] was calculated to assess the strength of the linear relationship between the two values.

### 5.2 Clinical validation: association with clinical phenotypes

The test set comprised 196 cases, including 133 patients of group A (more than 50% stenosis), 30 cases of group B (no obstructive CAD with positive functional testing), and 33 group C (less than 50% stenosis without CMVD). The reference standard for CMVD was established based on typical angina symptoms, exclusion of obstructive CAD on angiography, and abnormal TIMI flow grad (less than grad 3), reflecting routine clinical practice.

### 5.2.1 Inter-group comparison of angiographic parameters

Following the initial stenosis detection step, inter-group comparisons were performed exclusively between the CMVD group and the normal group. Intergroup comparisons were performed for DL-TMPFC values and conventional angiographic parameters (TIMI flow grade[18], TIMI frame count (TFC)[19]). Due to the retrospective nature of this study, invasive microvascular indices (e.g., IMR, CFR) were not available. Therefore, TIMI flow grade and TFC were used as comparators, acknowledging their limited capacity to reflect microvascular function.

### 5.2.2 Diagnostic threshold of DL-TMPFC

The optimal diagnostic threshold of DL-TMPFC for recognizing CMVD was first determined using receiver operating characteristic (ROC) curve[20] analysis in the training set (include group B and group C), maximizing the Youden index (sensitivity + specificity -1). This threshold was then applied to the test set (include group B and C) to classify patients as suspected CMVD or not. Classification performance was assessed against the true CMVD diagnosis by calculating area under the curve (AUC), sensitivity, specificity, positive predictive value, and negative predictive value.

## 5.3 Incremental value assessment

Here we designed the experiments of correlation analysis and risk stratification to evaluate the incremental quantitative analysis value of the proposed DL-TMPFC.

### 5.3.1 DL-TMPFC correlation with CMVD severity

Pearson's correlation coefficients[17] were calculated in test cases of group B between DL-TMPFC values and four clinical parameters that indirectly reflecting CMVD severity. For left coronary artery myocardial ischemia, we used the mitral E/A ratio (mitral valve early (E) to late (A) diastolic filling velocity ratio) and tissue Doppler $e'/a'$ ratio (mitral annular early ($e'$) to late ($a'$) diastolic velocity ratio), both obtained from preprocedural echocardiography. Graphically, the relationship between DL-TMPFC and each severity indicator was visualized using scatter plots with fitted regression lines and 95% confidence intervals, with correlation coefficients ($r$) and $p$ values annotated in the plots.

### 5.3.2 Risk stratification by DL-TMPFC

Test patients of group B were stratified into three groups according to DL-TMPFC values: low TMPFC group, intermediate TMPFC group, and high TMPFC group. The above three parameters were compared across the three groups using the Jonckheere-Terpstra test[21] for trend to evaluate dose-dependent relationships. To illustrate the stratification effect, box plots were generated to display the distribution of the four parameters across the three TMPFC groups, with p for trend values and group-wise comparisons annotated.

# Results

## 1. Coronary stenosis detection network development

The developed coronary stenosis detection model demonstrated exceptional performance in real-time coronary stenosis detection across 196 test cases. Quantitative evaluation revealed outstanding detection accuracy, with mAP50 reaching 0.991, indicating near-flawless stenosis identification. The mAP50-95 of 0.787 suggested precise boundary localization essential for clinical decision making. Notably, the high recall of 0.970 demonstrated the model's sensitivity in capturing true stenotic lesions, while the precision of 0.957 ensured a low false positive rate, both critical for safe and effective clinical deployment. In terms of computational efficiency, the proposed network achieved 59.5 GFLOPs with 19.58 million parameters, demonstrating a favorable balance between model compl complexity and detection accuracy. These results validated the network as a highly accurate and reliable component for the initial identification of coronary stenosis within the integrated DL framework.

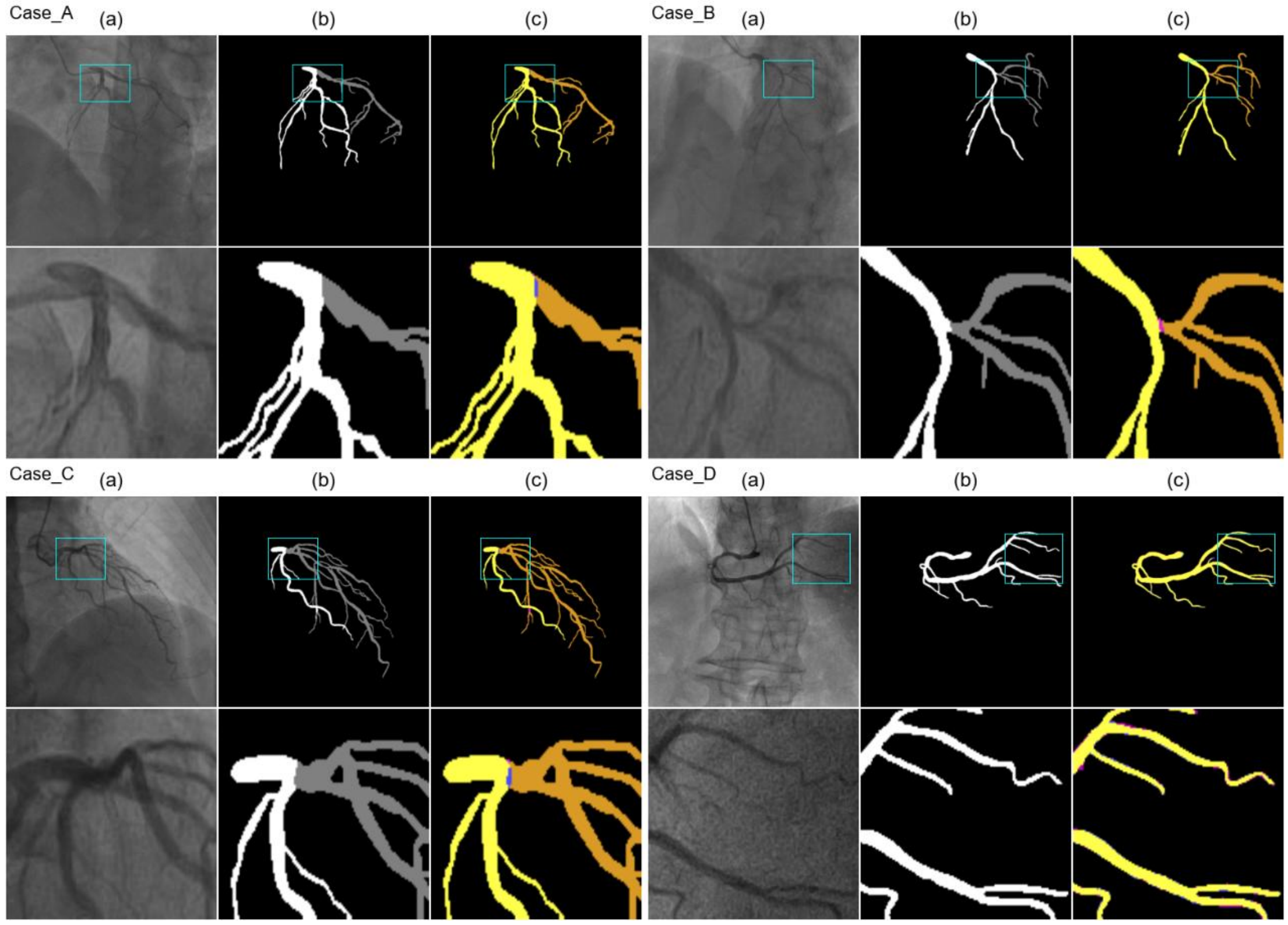

**Figure 4.** Representative examples of the coronary segmentation results.

## 2. Coronary multi-class territory-aware segmentation network development

The segmentation network achieved exceptional accuracy on the 196 hold-out test cases. Qualitative assessment across all the test datasets, mean Likert score was 4.5 with 82% rated more than 4. Qualitative performance across LAD, LCX and RCA revealed mean Likert scores of 4.4 for LAD territories, 4.5 for LCX territories, and 4.6 for RCA territories.

Figure 4 plotted the representative examples of the coronary segmentation results. Cases A, B and C illustrated the left coronary artery, and case D illustrated the right coronary. For each case, (a) depicted the raw angiogram, (b) showed the ground-truth label (bright: LAD; dark: LCX), and (c) displayed the network segmentation results (yellow: LAD); orange: LCX). Magenta indicated false positives, while blue indicated false negatives. The left coronary arteries were accurately segmented into left anterior descending (LAD) and left circumflex (LCX) arteries with clear delineation of their respective anatomical distributions. The right coronary artery (RCA) also demonstrated high-fidelity segmentation throughout its course. The segmentation consistently achieved high specificity across all the territories by effectively excluding non-coronary structures (diaphragm, spine). Vessel continuity was robust at bifurcation cores.

Quantitative evaluation over all the test datasets demonstrated a mean Dice Similarity Coefficient (DSC) of 91.84%, mean sensitivity of 92.02%, mean 95% Hausdorff Distance of 4.2074 mm, mean Intersection over Union (IoU) of 85.02%. The high segmentation fidelity DSC coupled with exceptional sensitivity confirmed the algorithm's capability to accurately delineate coronary anatomy. Crucially, the submillimeter of HD95 indicated precise boundary localization, meeting the clinical tolerance for stenosis quantification. The robust IoU reflected effective vessel extraction.

## 3. TMPFC measurement algorithm

The automated identification of the initial and final frames for TMPFC calculation relied on adaptive parameters $(N_1, \delta_1, N_2, \delta_2)$, whose optimal values were empirically derived from the training dataset. The median filter window size for $F_1, F_2$ was 4% of the sequence length, with a maximum of 3 and a maximum of 11 (odd numbers only). $N_1$ for $F_1$ was 12 frames width with a quantile of 0.3. $N_2$ for $F_2$ was 5 frames width with a quantile of 0.3. $\delta_1, \delta_2$ were 0.9 and 0.1 respectively.

The robustness of the selected parameters was illustrated in Figure 5, which plotted the temporal dynamics for three representative vascular segments sequences. The x-axis represented the frame indices, and the y-axis represented $A_t$ value (segmented pixels count). The y-coordinates (identifying first and last frames) were algorithmically determined by the computed adaptive thresholds, while the corresponding x-coordinates (frame indices) were derived from these vessel-specific curves. The vertical dashed lines in Figure 5, indicating the algorithmically determined frames, demonstrated that $F_1$ was triggered at the onset of the plateau phase in the curve. This critical point corresponded to the moment that the myocardial perfusion territory first became fully opacified, satisfying the standard TMPFC definition for the initial frame $F_1$. Conversely, $F_2$ was reliably activated during the terminal phase of contrast clearance, accurately identifying the point of 'near-complete washout'. The close agreement between the automated frame selections and the characteristic phases of the perfusion curves confirmed the precision and physiological validity of the optimized threshold parameters.

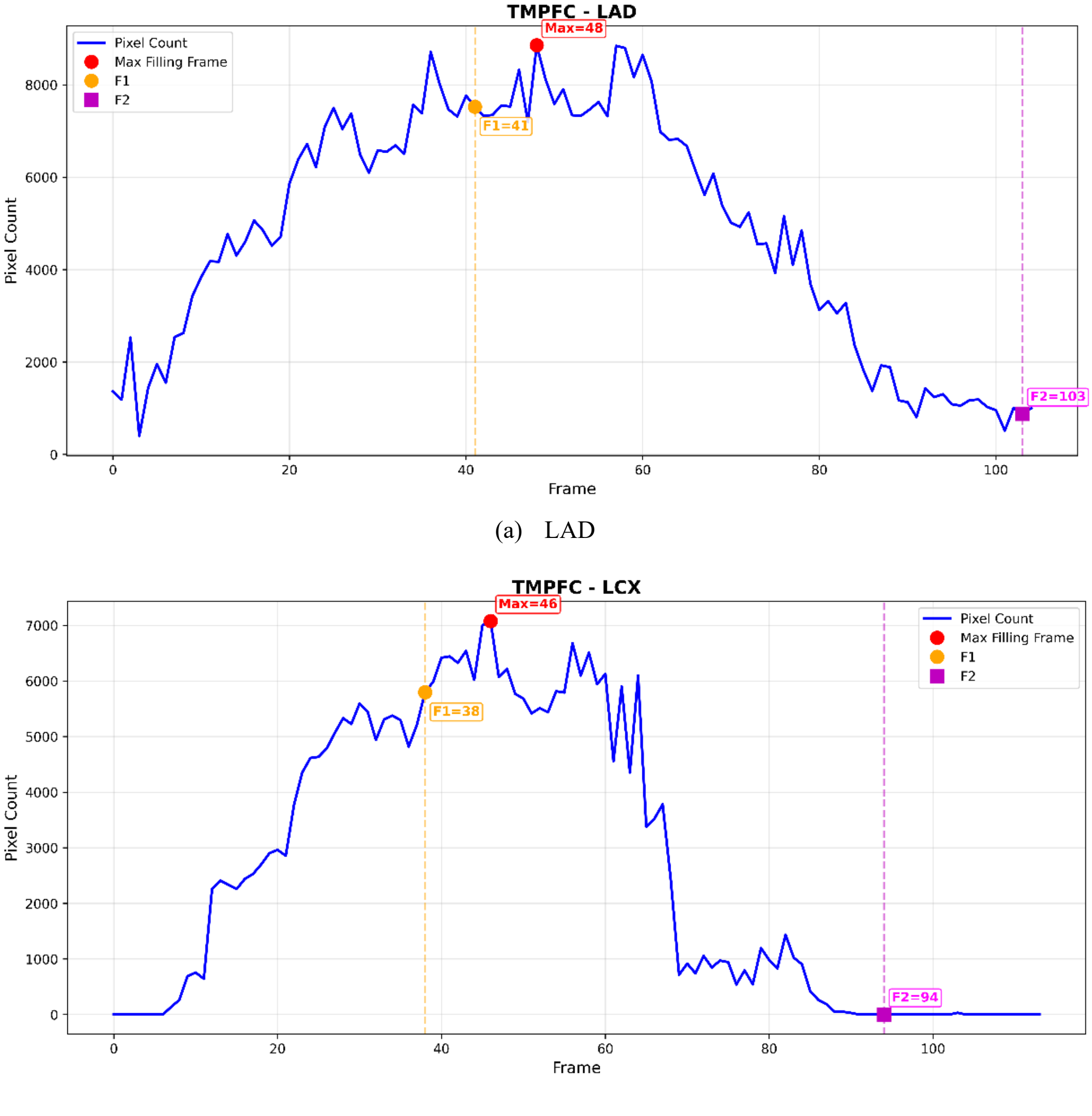

(a) LAD

(b) LCX

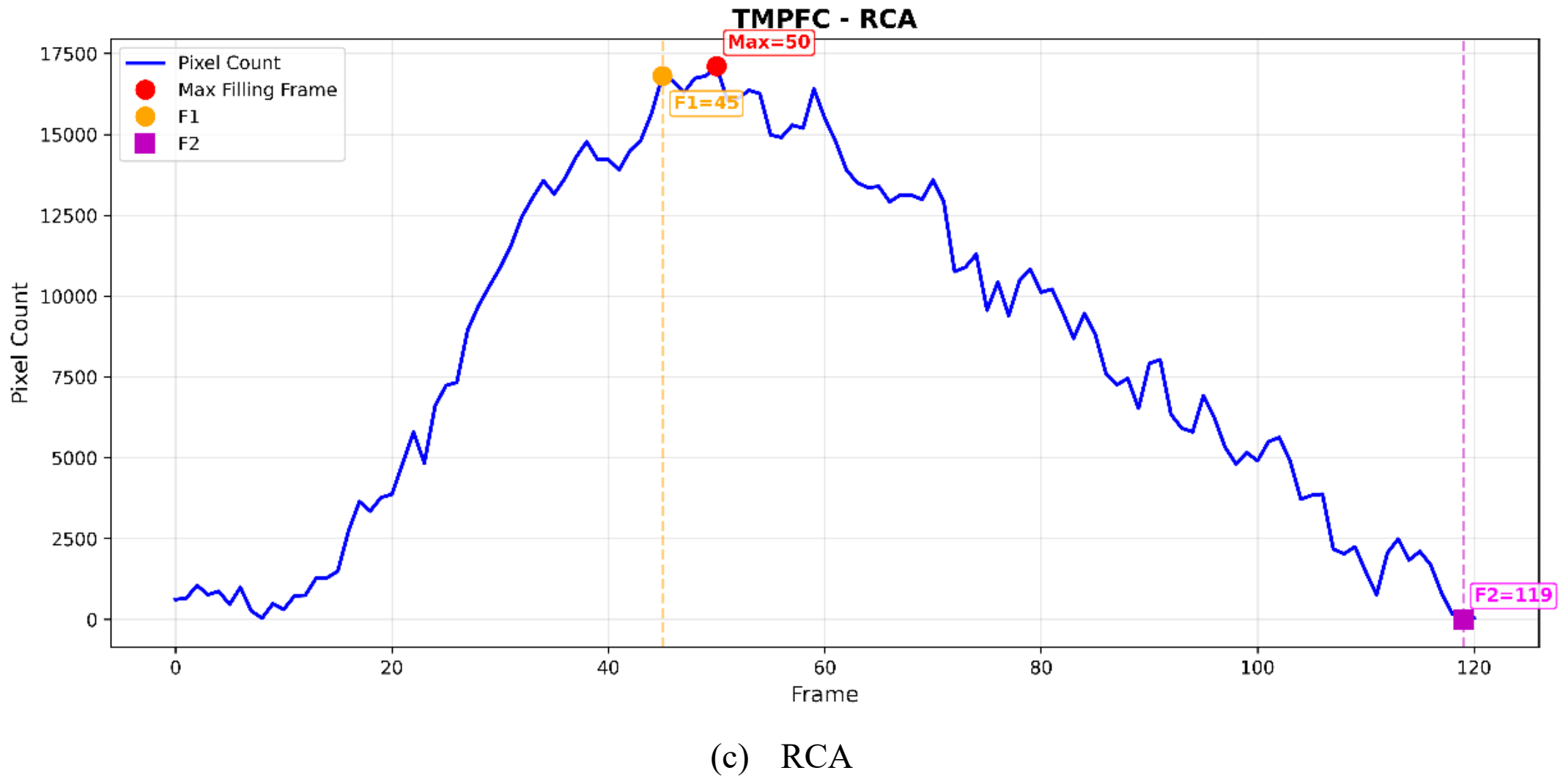

(c) RCA

**Figure 5.** Validation of TMPFC calculation parameters selection: the correlation between segmented pixel counts and frame indices (fps=15).

## 4. Comprehensive performance validation

### 4.1 Technical validation: agreement versus manual TMPFC

The proposed DL-TMPFC framework demonstrated excellent agreement with expert manual measurements. As detailed in Figure 6(a), the Bland-Altman analysis revealed a mean bias of -0.93 frames, with 95% limits of agreement (LoA) ranging from -5.33 to +3.47 frames. The strong linear relationship between the two methods was confirmed by a Pearson's correlation coefficient of $r$ =0.989 ($p$ <0.001), as visually supported by the scatter plot in Figure 6(b).

### 4.2 Clinical validation: association with clinical phenotypes

#### 4.2.1 Inter-group comparison of angiographic frames

Inter-group comparisons were performed exclusively between the group B (confirmed CMVD cases) and group C (control group). DL-TMPFC values were significantly higher in group B than in group C (median [IQR]: 117.5 [110.5-132.5] frames vs. 60 [54-66] frames, $p$ < 0.001). As expected, all patients of group B had TIMI flow grade less than 2, whereas all patients of group C had grade 3 ($p$ < 0.001). Corrected TFC value showed a trend toward higher values in the CMVD group but did not reach statistical significance (32.4±8.1 vs. 28.6±6.3 frames, p = 0.08).

#### 4.2.2 Diagnostic threshold of DL-TMPFC

The optimal diagnostic threshold for DL-TMPFC was first determined in the mixed training dataset (cases from group B and group C) by maximizing the Youden index, yielding a threshold of 87 frames. When this threshold was applied to the mixed test dataset, DL-TMPFC achieved an area under the curve (AUC) of 0.985 for detecting CMVD. The sensitivity was 98.3% (95% CI: 91.0-99.9%), specificity was 97.2% (95% CI: 90.0-99.5%), positive predictive value was 96.7% (95% CI: 88.5-99.6%), and negative predictive value was 98.6% (95% CI: 92.5-99.9%). These findings demonstrated that DL-TMPFC accurately identified CMVD cases from real-world angiographic data encompassing a full spectrum of coronary pathologies.

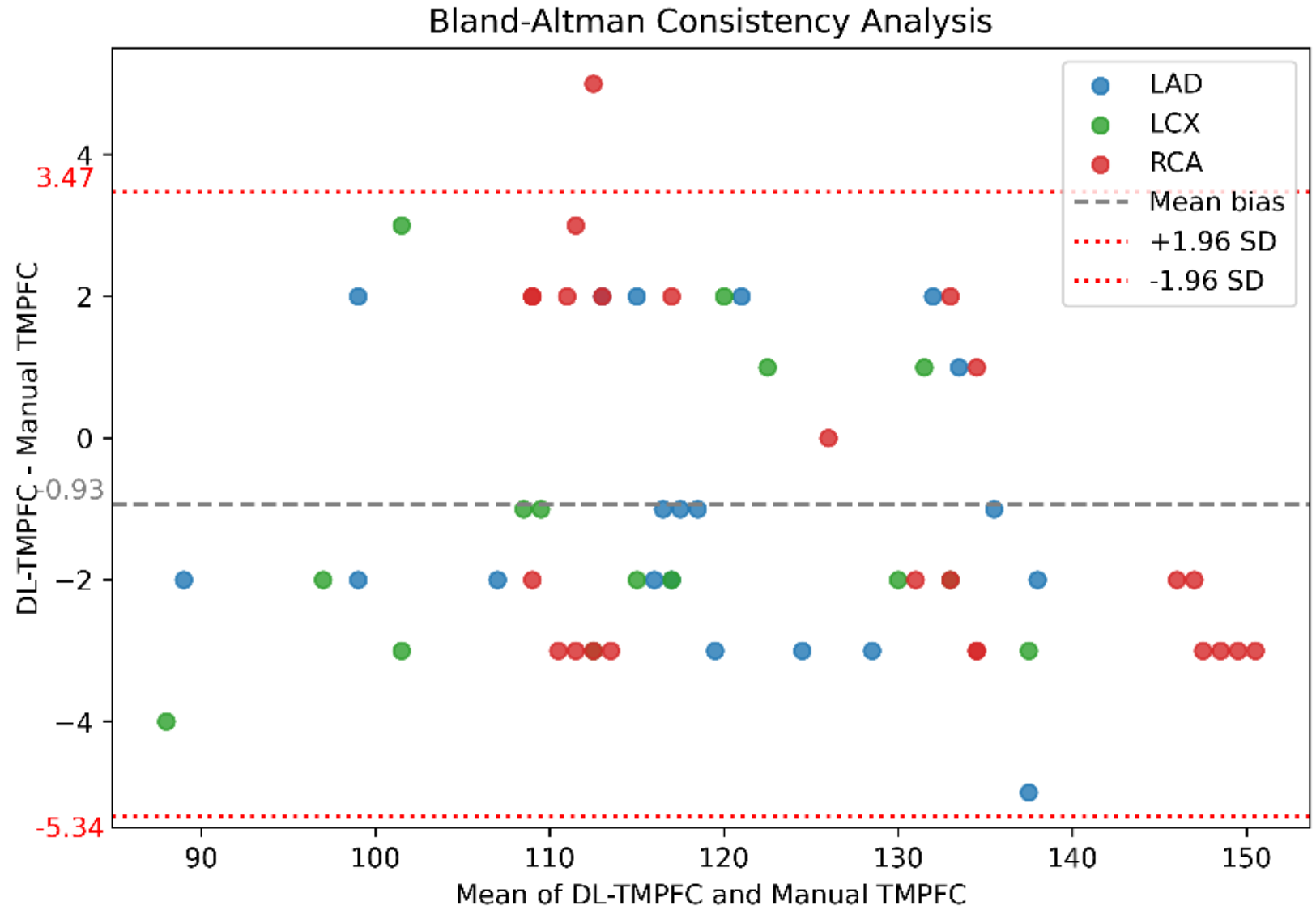

(a) Bland-Altman analysis plot.

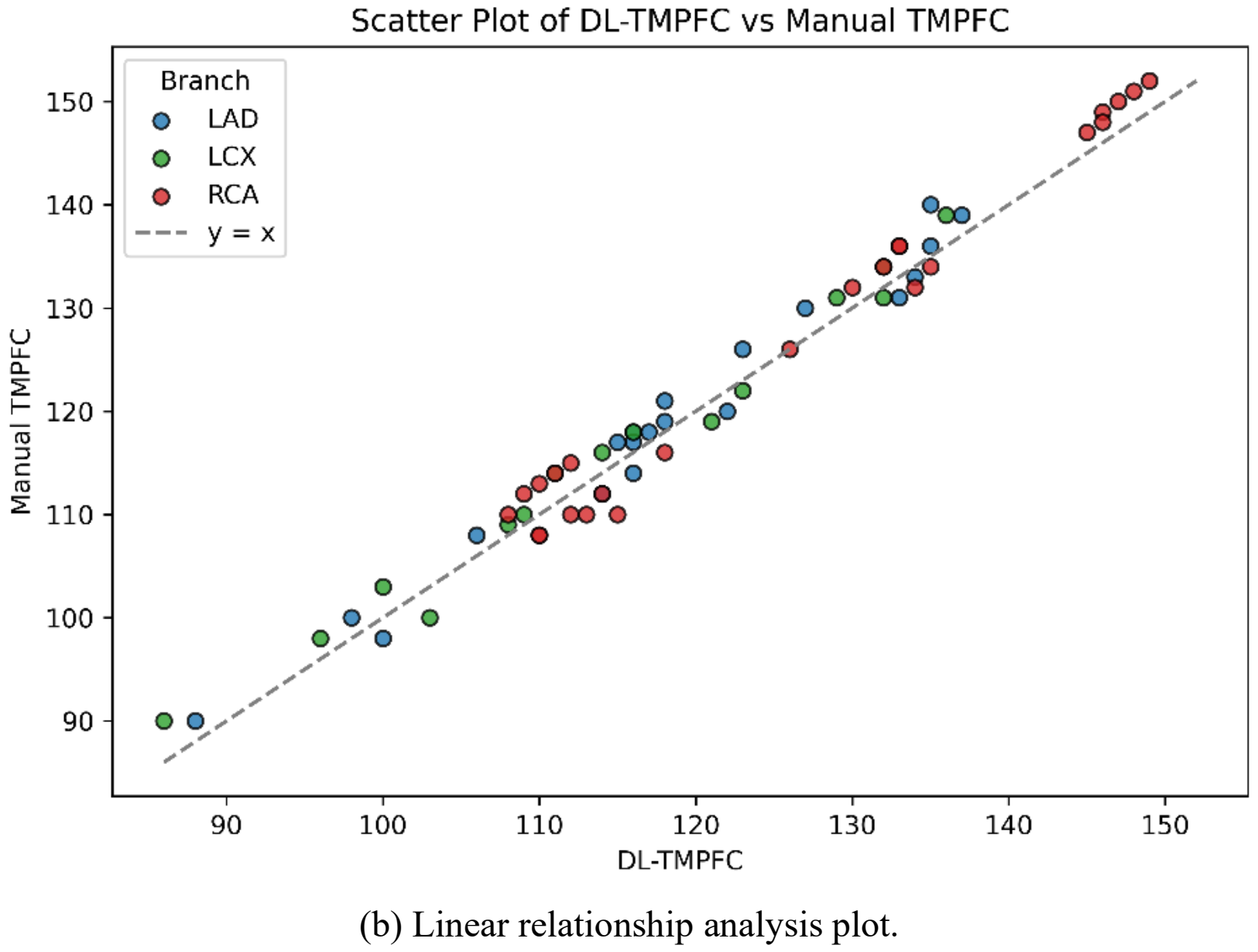

(b) Linear relationship analysis plot.

**Figure 6.** Agreement analysis between DL-TMPFC and manual TMPFC.

## 4.3 Incremental value assessment

### 4.3.1 DL-TMPFC correlation with CMVD severity

In test cases of group B, Pearson's correlation analysis revealed a strong positive correlation between DL-TMPFC values and E/A ratio ($r$ = -0.970, 95% CI: -0.985 to -0.940, $p$ < 0.001). Similarly, a significant positive correlation was observed between DL-TMPFC values and $e'/a'$ ratio ($r$ = -0.934, 95% CI: -0.967 to -0.873, p < 0.001). Scatter plots with fitted regression lines and 95% confidence intervals demonstrated a clear linear relationship between DL-TMPFC and both severity indicators, with higher DL-TMPFC values corresponding to lower E/A ratio and $e'/a'$ ratio (Figure 7). These findings indicated that DL-TMPFC effectively captured the continuum of CMVD severity, supporting its role as a quantitative marker beyond binary classification.

### 4.3.2 Risk stratification by DL-TMPFC

When test patients of group B were stratified by DL-TMPFC into three groups according to the diagnostic threshold. Low group was of 87 to 113 frames, intermediate group was of 114 to 123 frames, and high group was of more than 127 frames. A stepwise decrease in E/A ratio and $e'/a'$ ratio was observed across the low, intermediate and high TMPFC groups ($p < 0.001$ for both). Box plots revealed clear gradients in both severity measures across the three groups, with the higher TMPFC group showing the lower E/A and $e'/a'$ ratios (Figure 8). In group B, the clinical data showed E/A <1 and $e'/a' < 1$. Within this range of impaired relaxation, lower E/A and $e'/a'$ ratios indicated more severe myocardial ischemia and indirectly corresponded to greater severity of coronary microvascular dysfunction (CMVD). These findings demonstrated that DL-TMPFC enabled quantitative risk stratification, capturing the full spectrum of disease severity in patients with coronary microvascular dysfunction.

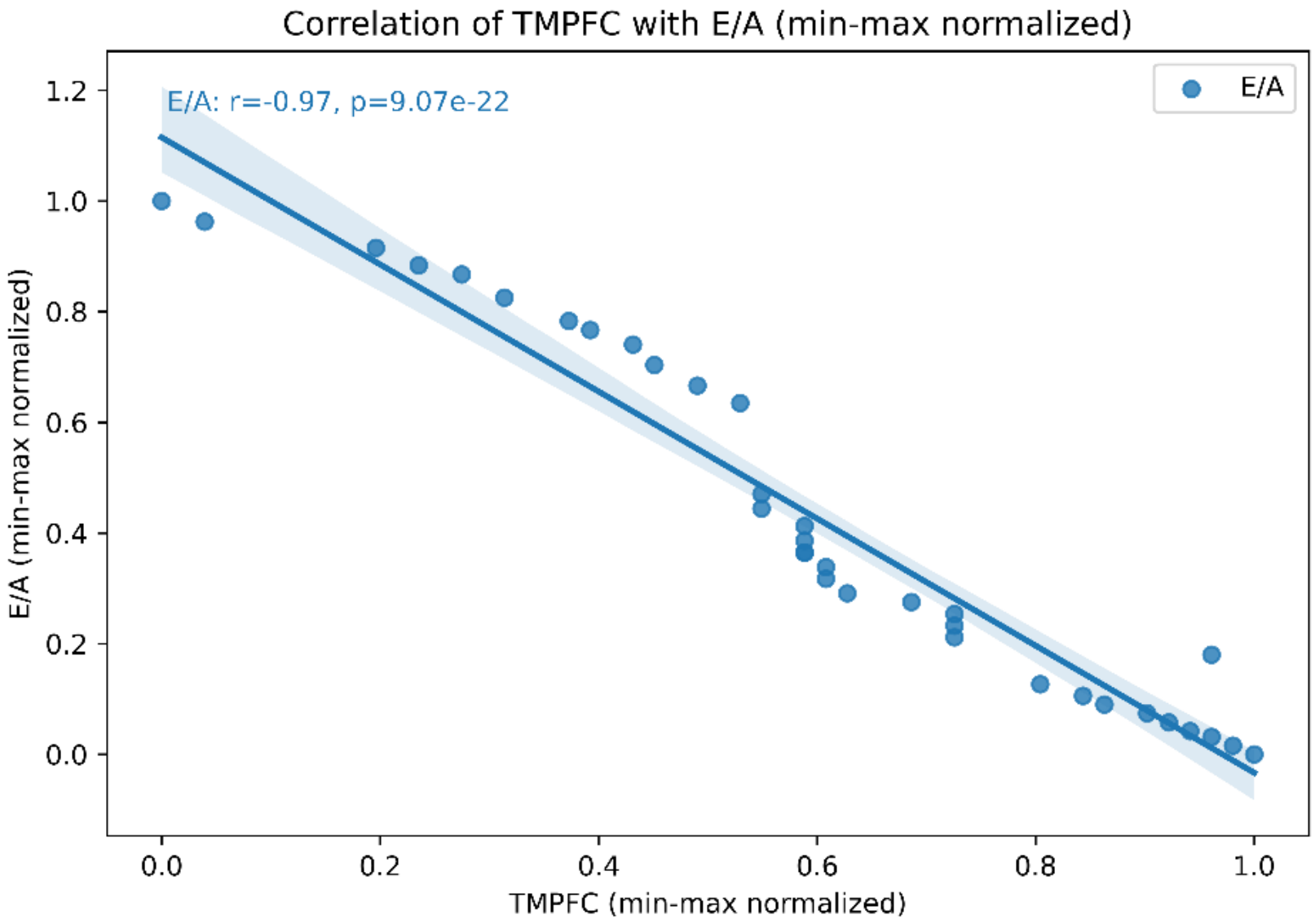

(a) Scatter plots of DL-TMPFC versus E/A ratio (normalized value).

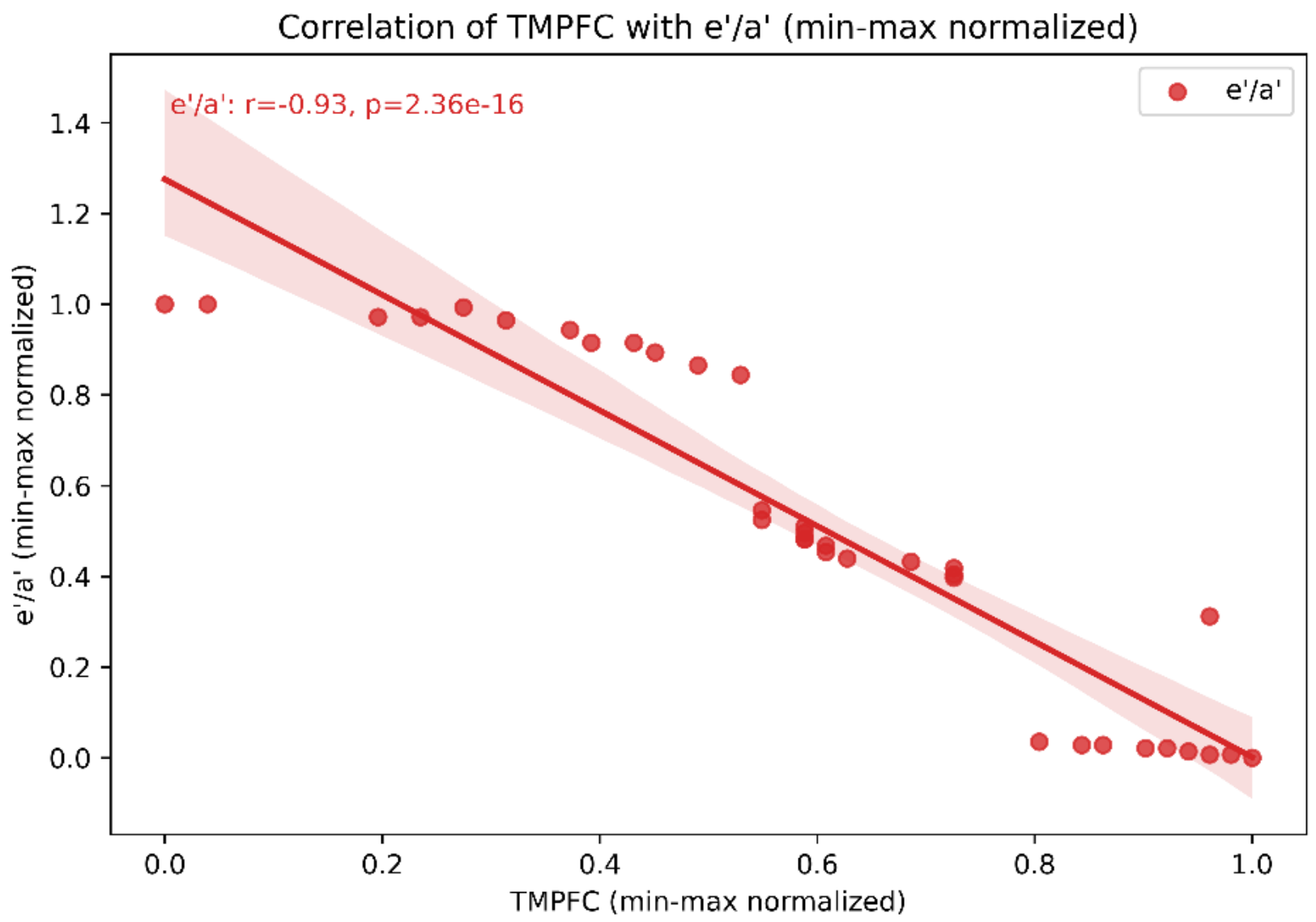

(b) Scatter plots of DL-TMPFC versus $e'/a'$ ratio (normalized value).

**Figure 7.** DL-TMPFC correlation with CMVD severity.

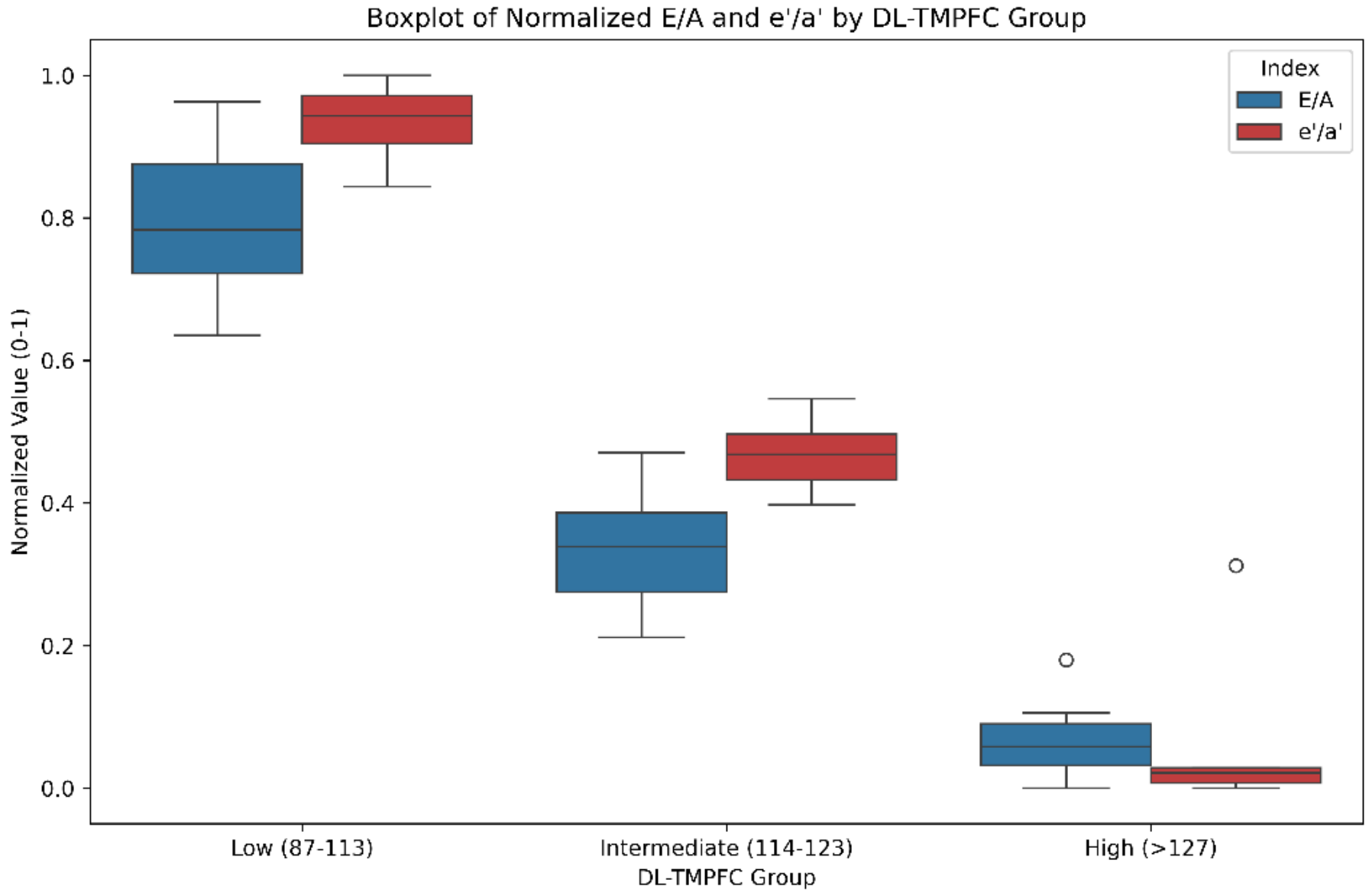

**Figure 8.** Risk stratification of CMVD severity by DL-TMPFC tertiles.

# Discussion

The proposed DL-TMPFC framework has several limitations that warrant consideration. First, the study faces constraints in clinical validation. Although the technical efficacy of DL-TMPFC was established, its prognostic value for predicting major adverse cardiovascular events (MACE) remains unverified, as the original TMPFC studies evaluated MACE in acute myocardial infarction cohorts, whereas our dataset comprised predominantly patients with stable coronary artery disease. Second, this method requires high-quality angiographic sequences with optimal opacification and full coverage of the cardiac cycle from initial filling to complete washout without premature contrast fading.

# Conclusions and Clinical Implications

The proposed DL-TMPFC method enables automated, real-time quantification of CMVD, eliminating observer variability and streamlining workflows. The method demonstrates robust performance in identifying microvascular dysfunction, particularly in high-risk populations with non-obstructive coronary artery disease, offering rapid screening potential.

For CMVD patients with angina, guideline-directed medical therapy remains the cornerstone of management. Such therapy includes $\beta$-blockers, calcium channel blockers, nicorandil, ivabradine, ranolazine, and angiotensin-converting enzyme inhibitors (ACEI). The automated TMPFC enables objective pre- and post-treatment comparison, offering a robust imaging biomarker for therapeutic efficacy. Although repeated invasive angiography is infrequent in clinical routine, retrospective analysis of existing angiographic data could further validate its utility.

Procedure-related CMVD is a recognized complication following percutaneous coronary intervention (PCI). The DL-TMPFC framework provides real-time, automated feedback on microvascular function during the procedure, potentially guiding interventional strategies to mitigate iatrogenic microvascular injury. These applications underscore the translational potential of AI-enabled TMPFC quantification as both a diagnostic and therapeutic monitoring tool in evolving cardiovascular practice.

# Declarations

**1. Consent**

Written informed consent was obtained from all individual participants included in the study.

### 2. Data availability statement

The datasets and software code are available in the github repository, [https://github.com/DrThink-ai/DL-TMPFC].

### 3. Conflict of interest

None declared.

### 4. Author contributions

SL: Data curation, Investigation, Methodology, Software, Validation, Visualization, Writing-original draft. YQH: Data curation, Software, Formal analysis, Resources. CKH, XGG, HCT: Conceptualization, Data curation, Formal analysis, Project administration, Resources, Writing-review & editing. CYK, XZ, LH, JYZ, SX: Data curation, Investigation, Visualization.

## Figure legends

Figure 1. Schematic of the proposed DL-TMPFC framework.

Figure 2. Detailed description of the development of coronary stenosis detection network model.

Figure 3. Detailed description of the development of DL-driven TMPFC measurement method.

Figure 4. Representative examples of the coronary segmentation results.

Figure 5. Validation of threshold parameters selection: the correlation between segmented pixel counts and frame indices.

Figure 6. Agreement analysis between DL-TMPFC and manual TMPFC.

Figure 7. DL-TMPFC correlation with CMVD severity.

Figure 8. Risk stratification of CMVD severity by DL-TMPFC tertiles.